\documentclass[10pt,twocolumn,letterpaper]{article}

\usepackage[pagenumbers]{cvpr} 

\usepackage{graphicx}
\usepackage{amsmath}
\usepackage{amssymb}
\usepackage{booktabs}
\usepackage{tablefootnote}
\usepackage[flushleft]{threeparttable} 

\usepackage[pagebackref,breaklinks,colorlinks]{hyperref}

\usepackage[capitalize]{cleveref}
\crefname{section}{Sec.}{Secs.}
\Crefname{section}{Section}{Sections}
\Crefname{table}{Table}{Tables}
\crefname{table}{Tab.}{Tabs.}

\def\cvprPaperID{*****} 
\def\confName{CVPR}
\def\confYear{2023}

\begin{document}

\title{Enhancing Transformer Backbone for Egocentric Video Action Segmentation}
\author{Sakib Reza\textsuperscript{1, 2}, Balaji Sundareshan\textsuperscript{1}, Mohsen Moghaddam\textsuperscript{1, 2}, Octavia Camps\textsuperscript{1} \\
\textsuperscript{1}College of Engineering, Northeastern University, Boston, MA\\
\textsuperscript{2}Khoury College of Computer Sciences, Northeastern University, Boston, MA\\
{\tt\small \{reza.s, sundareshan.b, mohsen, o.camps\}@northeastern.edu}
}
\maketitle

\begin{abstract}
   Egocentric temporal action segmentation in videos is a crucial task in computer vision with applications in various fields such as mixed reality, human behavior analysis, and robotics. Although recent research has utilized advanced visual-language frameworks, transformers remain the backbone of action segmentation models. Therefore, it is necessary to improve transformers to enhance the robustness of action segmentation models. In this work, we propose two novel ideas to enhance the state-of-the-art transformer for action segmentation. First, we introduce a dual dilated attention mechanism to adaptively capture hierarchical representations in both local-to-global and global-to-local contexts. Second, we incorporate cross-connections between the encoder and decoder blocks to prevent the loss of local context by the decoder. We also utilize state-of-the-art visual-language representation learning techniques to extract richer and more compact features for our transformer. Our proposed approach outperforms other state-of-the-art methods on the Georgia Tech Egocentric Activities (GTEA) and HOI4D Office Tools datasets, and we validate our introduced components with ablation studies. The source code and supplementary materials are publicly available on \url{https://www.sail-nu.com/dxformer}. \footnote{This work is supported by the NSF Grant No. 2128743. Any opinions, findings, and conclusions expressed in this material  are those of the authors and do not necessarily reflect the views of the NSF.}
\end{abstract}

\section{Introduction}
\label{sec:intro}
Automated detection and segmentation of human activities from an egocentric perspective have numerous applications in fields such as mixed reality, human behavior analysis, and robotics \cite{lea2017temporal,kwon2022context}. However, egocentric action segmentation is particularly challenging due to several factors. First, the videos are untrimmed and can span several minutes, making it difficult to assign an action label to each frame accurately. Second, egocentric videos often have occlusions, where the camera wearer's body or objects in the foreground obstruct the view of the action. Random movement and camera motion can also lead to inconsistent viewpoints, making it challenging to track actions across frames. Third, active hand-object interactions, which are prevalent in egocentric videos, can result in actions slightly outside the frame, further complicating the segmentation task. To address these challenges, action segmentation methods for egocentric videos focus on modeling the temporal relations among frames using pre-extracted frame-wise feature sequences, while also considering the unique characteristics of egocentric videos.

Recent advances in action segmentation models have shown promising results through the use of cutting-edge visual language feature representations \cite{li2022bridge, wang2021actionclip}. However, there remains scope for improvement in the transformer backbone, which serves as the fundamental component of these models. Specifically, enhancing the transformer backbone can aid in precise action identification by effectively attending to salient regions within the video. A strengthened attention mechanism within the improved transformer plays a pivotal role in capturing temporal dependencies, ultimately leading to improved model accuracy.

In this paper, we introduce DXFormer, a new transformer-based architecture that improves the state-of-the-art transformer backbone for action segmentation by introducing a dual dilated attention mechanism and cross-connections between encoder and decoder blocks. In addition, we build on advanced visual-language representation learning approaches to extract richer and more compact characteristics for the transformer. We evaluate DXFormer on two challenging egocentric video datasets and show that it outperforms other state-of-the-art action segmentation methods both quantitatively and qualitatively. Furthermore, we conduct an ablation study to validate the effectiveness of newly added components in improving the results with respect to various metrics.

\begin{figure*}[htbp]
	\centerline{\includegraphics[width=0.87\linewidth]{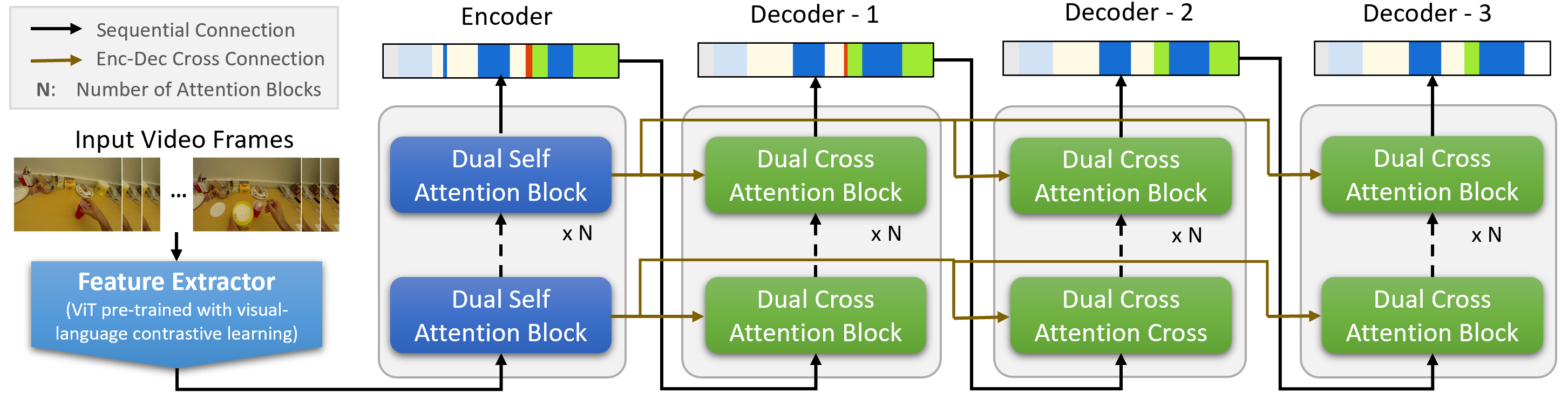}}
    \caption{The proposed DXFormer model for temporal action segmentation uses a multi-decoder approach  to capture  temporal dependencies  more effectively.} \label{dxformer}
    \vspace{-4mm}
\end{figure*}

\section{Method}
The proposed DXFormer is an improved backbone model for action segmentation, based on the ASFormer \cite{chinayi_ASformer} baseline model. It introduces two innovative techniques, dual dilated attention (DA) and encoder-decoder cross-connection (CC), to elevate its performance. The overall architecture of DXFormer, depicted in Figure \ref{dxformer}, is presented to provide a high-level overview. The proposed model comprises a single encoder and three decoders. Initially, video frames are passed through a feature extractor model, which is a vision transformer pre-trained using visual language contrastive learning methods (e.g., BridgePrompt \cite{li2022bridge} and CLIP \cite{radford2021learning}). The encoder leverages the extracted frame-wise features to generate initial action predictions for each frame. Subsequently, the three decoders refine these initial outputs, yielding improved frame action labels as the final output. This multi-decoder approach effectively enhances the accuracy of the action segmentation by capturing temporal dependencies in the egocentric video data.


\subsection{Dual Dilated Attention Mechanism}

\begin{figure}[t]
	\centerline{\includegraphics[width=0.77\linewidth]{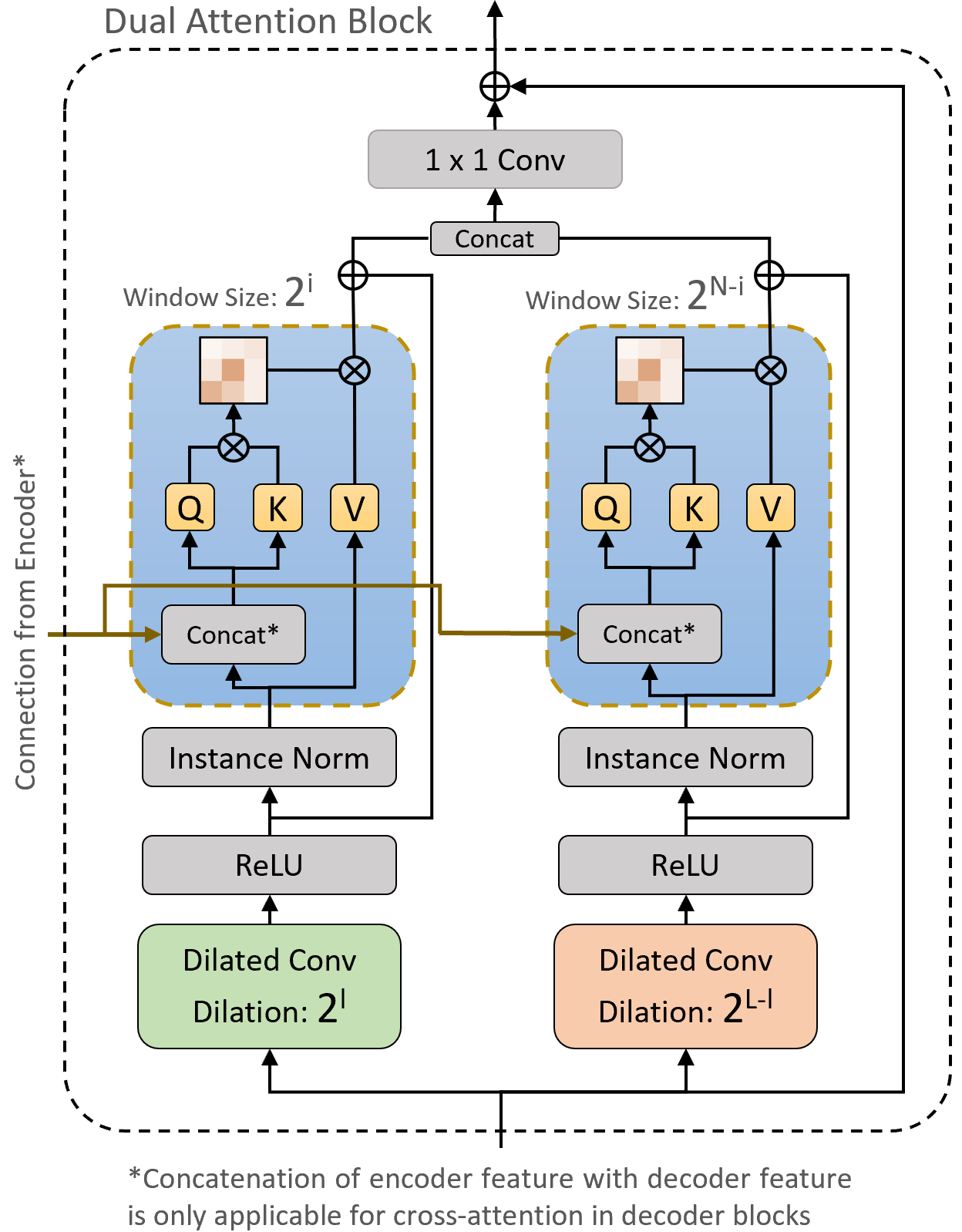}}
    \caption{Dual Dilated Attention (DA) block.} \label{da}
    \vspace{-6mm}
\end{figure}

 The baseline model, ASFormer \cite{chinayi_ASformer}, maintains a hierarchical pattern representation to focus first on local information and then gradually expand the attention span to capture the global feature. In this approach, while the attention span is very large for the higher attention blocks, lower blocks may suffer from a small attention span. Due to the smaller attention windows, the initial attention blocks capture the local context from neighboring frames quite well, but miss the global contexts. On the other hand, the higher-attention blocks capture the global context quite well because of the larger attention span, but they miss out on the local contexts. To address the limitations mentioned above, we propose a dual-dilated attention (DA) approach, to help the model to capture both global and local contexts adaptively in both lower and higher attention blocks. Figure \ref{da} shows the architecture of the proposed DA module. 

The new DA module consists of two different attention branches, one with an increasing window size and the other with a decreasing window size. Each attention branch starts with a dilated convolution layer followed by an attention layer. Following \cite{chinayi_ASformer}, we keep the dilation size of the convolution layers and the window size of its attention layer the same. For the first attention branch, the window size is doubled at each block (e.g., $2^{i}$, where $i$ = 1, 2,...). On the other hand, for the second branch, the window size is halved in each block (e.g., $2^{N-i}$, where $N$ = 9 and $i$ = 1, 2,...). Hence, in the initial attention blocks, the first branch has a small attention span and captures local context from neighboring frames. In contrast, the second branch starts with a large attention span and captures global context exploring both near and distant frames. These two branches are then merged through concatenation, followed by a convolution operation. This adaptive combination enables the model to dynamically learn the emphasis placed on each branch, facilitating the effective capture of both local and global contexts.   

The proposed DA uses different attention mechanisms for the encoder and the decoders. The encoder uses the self-attention mechanism and the output of the previous layer as the query $Q$, the key $K$, and the value $V$. On the other hand, the decoders use a cross-attention mechanism where the query $Q$ and the key $K$ are generated by concatenating the outputs from its previous layer and the corresponding encoder attention block, and the value $V$ is taken from the previous layer as self-attention. 

\subsection{Encoder-Decoder Cross Connections}
Connecting the encoder and decoder sequentially in the transformer backbone can lead to a semantic gap between their features. For example, in the ASFormer \cite{chinayi_ASformer}, the first decoder takes the output from the last block of the encoder as input, while the subsequent decoders take the output from their previous decoder's last block as input. Although the output of any encoder or decoder's last block already contains global context of the input, sharing only this feature with the decoder blocks can result in a loss of local information. This can hinder the learning process of both local and global patterns using DA and limit the effectiveness of the overall architecture. 

To address the issue of losing local context, we propose cross-connections between the encoder and decoder blocks. These cross-connections connect each decoder block with its corresponding encoder block (as depicted in Figure \ref{dxformer}) and pass the local information from the lower encoder block to its corresponding decoder blocks, preventing the model from losing important local information at the decoders. Each cross-connection takes the output of an encoder block and passes to its corresponding decoder block. In the decoder blocks, the features are concatenated with the input of their cross-attention layer, and the result is used as the query $Q$ and the value $V$, as shown in Figure \ref{da}. 

\section{Experiments}

\subsection{Datasets and Experiment Setup}
We utilize two challenging egocentric video datasets to assess the performance of the proposed approach. The first dataset, \textbf{Georgia Tech Egocentric Activities (GTEA)} \cite{fathi2011learning}, comprises 28 instructional videos recorded from an egocentric perspective. It includes 11 distinct action classes representing daily kitchen activities. The second dataset, \textbf{HOI4D Office Tools}, is a subset of the larger HOI4D dataset \cite{liu2022hoi4d} and consists of 553 egocentric videos depicting hand interactions with office tools such as scissors, pliers, and staplers. It covers 12 action classes related to office tool usage. To evaluate the performance of our approach, we conducted a four-fold cross-subject validation for both datasets. This evaluation technique ensures robustness and mitigates the influence of subject-specific biases on the results.

For our experiments, we closely follow the settings employed by the ASFormer model \cite{chinayi_ASformer}, our baseline. However, for the large HOI4D dataset, we make specific adjustments for efficient training. We utilized a batch size of $8$, a learning rate of $0.001$, and incorporated $7$ attention blocks in each encoder and decoder.


\subsection{Effect of Dual-Attention and Cross-Connection}
To verify the influence of the new components in our DXFormer model, we perform an ablation study on the GTEA dataset. We specifically test various combinations of these elements and assessed how they affect the performance indicators. The addition of dual dilated attention (DA) increases the F1@50 score by 1.2\% and accuracy by 1\%, according to the findings of the ablation study. The best F1@10 score is achieved by including cross-connections (CC) between the encoder and decoder blocks. Finally, adding both DA and CC allows us to achieve the highest edit score. The findings in Table \ref{Tab:1} show how well the newly incorporated components work and how they help the action segmentation models function better.  

\begin{table}
  \centering
  \caption{Performance comparison of different configurations of our proposed method vs. the baseline (ASFormer) on the GTEA dataset (using only BrPrompt feature representation). }
  \begin{threeparttable}
  \resizebox{0.9\linewidth}{!}
{
  \begin{tabular}{l | c c c |c | c} 
 \toprule
  \textbf{Backbone Model} & \multicolumn{3}{c|}{\textbf{F1}@\{10, 25, 50\}} & \textbf{Edit} & \textbf{Acc} \\ [0.5ex] 
 \hline
    Baseline & 91.4 & 89.9 & 81.4 & 88.2 & 80.7\\ 
    Baseline + DA & \underline{91.7} & \textbf{90.5} & \textbf{82.6} & \underline{88.8} & \textbf{81.7} \\ 
    Baseline + CC & \textbf{91.8} & \underline{90.3} & 80.6 & 88.3 & 80.9 \\
    Baseline + DA + CC & 91.0 & 89.8 & \underline{81.9} & \textbf{89.0} & \underline{81.3} \\ 
\bottomrule
\end{tabular}

}

  \begin{tablenotes}
  \item[*] \footnotesize DA = Dual Dilated Attention, CC = Cross-Connection
  \end{tablenotes}
  \end{threeparttable}
\label{Tab:1}
\vspace{-3mm}
\end{table}

\subsection{Effect of Feature Representations}
We experiment with various feature representations of video frames as input to our transformer-based DXFormer model. The results on the GTEA dataset show that DXFormer performs best with BridgePrompt feature representation, trained using visual-language contrastive learning with custom prompts \cite{li2022bridge}. We also evaluate the performance of two commonly used pre-trained models, I3D \cite{carreira2017quo} and ViT\cite{wang2021actionclip}, which are pre-trained on the Kinetics400 dataset\cite{kay2017kinetics}. The results demonstrate that BridgePrompt achieves the highest performance, followed by I3D, while ViT exhibits comparatively lower performance. Detailed comparison of these results is provided in Table \ref{Tab:2}.

\begin{table}
  \centering
  \caption{Performance comparison of different feature representations with DXFormer on the GTEA dataset.}
  \begin{threeparttable}
  \resizebox{0.7\linewidth}{!}
{
  \begin{tabular}{l | c c c |c | c} 
 \toprule
  \textbf{Feature} & \multicolumn{3}{c|}{\textbf{F1}@\{10, 25, 50\}} & \textbf{Edit} & \textbf{Acc} \\ [0.5ex] 
 \hline
    I3D & 89.1 & 88.0 & 78.1 & 84.9 & 80.3 \\ 
    ViT & 85.3 & 83.2 & 72.6 & 83.4 & 74.1  \\ 
    BrPrompt & \textbf{91.4} & \textbf{89.8} & \textbf{81.9} & \textbf{89.0} & \textbf{81.3} \\ 
\bottomrule
\end{tabular}

}

  \end{threeparttable}
  \label{Tab:2}
  \vspace{-5mm}
\end{table}




\begin{figure*}[htbp]
	\centerline{\includegraphics[width=\linewidth]{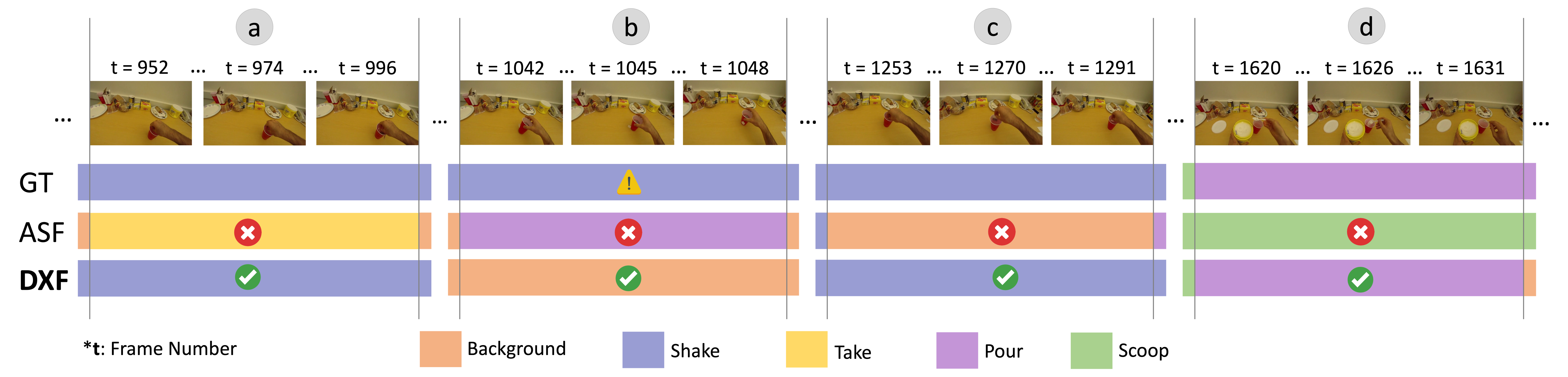}}
    \caption{Qualitative Evaluation: DXFormer outperforms the baseline (ASFormer) approach in challenging scenarios, as demonstrated on a representative video (\lq S1\_Tea\_C1\rq) from the GTEA dataset.} \label{qual}
    \vspace{-5mm}
\end{figure*}

\subsection{Comparison with the State-of-the-Art}
Table \ref{Tab:3} presents a comprehensive comparison between the proposed DXFormer model and other state-of-the-art approaches for temporal action segmentation on the GTEA dataset. Notably, our DXFormer model outperforms all other models, achieving state-of-the-art performance with an F1@50 score of 82.6\% and a frame-wise accuracy of 81.7\%. The incorporation of dual dilated attention and cross-connections in our architecture proves highly effective in improving the performance of the baseline. Moreover, on the HOI4D Office Tools dataset, where the action labels are more fine-grained, the cross-connection mechanism shows a particularly positive impact. In Table \ref{Tab:4}, DXFormer consistently outperforms other state-of-the-art models on this dataset as well. These results further validate the efficacy of the proposed approach, showcasing its potential to advance action segmentation and facilitate the development of robust video understanding models.

\begin{table}
  \centering
  \caption{Action segmentation results on GTEA dataset.}
  \begin{threeparttable}
  \resizebox{\linewidth}{!}
{
  \begin{tabular}{l | c c c |c | c} 
 \toprule
  \textbf{Method} & \multicolumn{3}{c|}{\textbf{F1}@\{10, 25, 50\}} & \textbf{Edit} & \textbf{Acc} \\ [0.5ex] 
 \hline
    MS-TCN++ \cite{li2020ms} & 88.8 & 85.7 & 76.0 & 83.5 & 80.1 \\
    BCN \cite{wang2020boundary} & 88.5 & 87.1 & 77.3 & 84.4 & 79.8 \\ 
    G2L \cite{gao2021global2local} & 89.9 & 87.3 & 75.8 & 84.6 & 78.5 \\ 
    ASRF \cite{ishikawa2021alleviating} & 89.4 & 87.8 & 79.8 & 83.7 & 77.3 \\
    SSTDA \cite{chen2020action} & 90.0 & 89.1 & 78.0 & 86.2 & 79.8 \\ 
    SSTDA+HASR \cite{ahn2021refining} & 90.9 & 88.6 & 76.4 & 87.5 & 78.7 \\ 
    \hline
    I3D+ASFormer \cite{chinayi_ASformer} \tnote{*} & 89.9 & 88.3 & 78.2 & 85.4 & 79.3 \\ 
    BrPrompt+ASFormer \cite{li2022bridge} \tnote{*} & \underline{91.4} & \underline{89.9} & 81.4 & 88.2 & 80.7 \\
    \hline
    BrPrompt+\textbf{DXFormer} (DA) & \textbf{91.7} & \textbf{90.5} & \textbf{82.6} & \underline{88.8} & \textbf{81.7}  \\ 
    BrPrompt+\textbf{DXFormer} & \underline{91.4} & 89.8 & \underline{81.9} & \textbf{89.0} & \underline{81.3} \\ 
\bottomrule
\end{tabular}

}

  \begin{tablenotes}
  \item[*] \footnotesize Reproduced results on our hardware configuration
  \item[] \footnotesize DA = Dual Dilated Attention Only
  \end{tablenotes}
  \end{threeparttable}
\label{Tab:3}
\vspace{-5mm}
\end{table}

\subsection{Qualitative Evaluation}
To further examine the performance of DXFormer, we conducted a visual analysis of the model outputs for several videos. Although DXFormer performs similarly to the baseline in most general cases, our findings indicate that it outperforms the baseline in challenging cases. For instance, in Segment (a) of Figure \ref{qual}, DXFormer accurately predicts the shaking action by capturing subtle local movements and global patterns in consecutive frames, while the baseline misclassifies it as a \lq take' action. In Segment (b), we discover an error in the ground truth labeling of an action as \lq shake', which should have been labeled as a \lq background'. Despite the incorrect ground truth label, our model correctly predicts the label, demonstrating reliability in subtle action scenarios. Segment (d) presents a case where the pouring action is followed by the scooping action in the video. The baseline ASFormer model misclassifies the pouring action as \lq scoop' due to the influence of the previous action, and irrelevant global patterns. However, DXFormer effectively captures more critical local contexts and correctly predicts the pouring action, demonstrating its ability to accurately segment actions in the presence of complex temporal dependencies. Overall, our model demonstrates greater flexibility and robustness by adaptively incorporating global and local context through the proposed dual dilated attention and cross-connection components, leading to better performance in challenging cases compared to baseline.

\begin{table}
  \centering
  \caption{Action segmentation results on HOI4D (Office Tools) dataset.}
  \begin{threeparttable}
  \resizebox{0.85\linewidth}{!}
{
  \begin{tabular}{l | c c c |c | c} 
 \toprule
  \textbf{Method*} & \multicolumn{3}{c|}{\textbf{F1}@\{10, 25, 50\}} & \textbf{Edit} & \textbf{Acc} \\ [0.5ex] 
 \hline
    MS-TCN \cite{farha2019ms} & 88.1 & 84.2 & 71.2 & 91.4 & 74.7 \\
    MS-TCN++ \cite{li2020ms} & 88.6 & 84.9 & 72.6 & 92.0 & 75.1 \\
    ASFormer \cite{chinayi_ASformer} & 89.4 & 85.6 & \underline{74.3} & \underline{93.0} & 76.1 \\ 
    \hline
    \textbf{DXFormer} (CC) & \textbf{89.8} & \textbf{86.0} & \textbf{74.7} & \textbf{93.4} & \underline{76.4}  \\ 
    \textbf{DXFormer} & \textbf{89.8} & \underline{85.9} & 73.8 & \underline{93.0} & \textbf{76.5} \\ 
\bottomrule
\end{tabular}

}

  \begin{tablenotes}
  \item[*] \footnotesize CLIP is used here for frame-wise feature extraction
  \item[] \footnotesize CC = Cross-Connection Only
  \end{tablenotes}
  \end{threeparttable}
\label{Tab:4}
\vspace{-5mm}
\end{table}

\section{Conclusion}
In this paper, we proposed two novel ideas, dual dilated attention and encoder-decoder cross-connection, to enhance the performance of the transformer backbones for the egocentric action segmentation task. We also conduct a comprehensive comparison of different feature representations for action segmentation models. Our proposed model demonstrates performance comparable to that of the baseline in general scenarios, while outperforming it in more challenging cases due to its ability to adaptively focus on temporally local and global contexts. Moving forward, we will continue to explore innovative ideas to further enhance the transformer backbone and extend our evaluations to a wider range of datasets, aiming to push the boundaries of egocentric action segmentation research.

{\small
\bibliographystyle{ieee_fullname}
\bibliography{CameraReady}
}

\end{document}